\documentclass{article}

\usepackage{microtype}
\usepackage{graphicx}
\usepackage{subfigure}
\usepackage{booktabs} 
\usepackage{amssymb}
\usepackage{natbib}

\usepackage{hyperref}

\usepackage[accepted]{icmlw2019generalization}

\icmltitlerunning{Adaptively Preconditioned Stochastic Gradient Langevin Dynamics}

\begin{document}

\twocolumn[
\icmltitle{Adaptively Preconditioned Stochastic Gradient Langevin Dynamics}




\begin{icmlauthorlist}
\icmlauthor{Chandrasekaran Anirudh Bhardwaj}{to}
\end{icmlauthorlist}

\icmlaffiliation{to}{Data Science Institute, Columbia University, New York, NY, USA}

\icmlcorrespondingauthor{Chandrasekaran Anirudh Bhardwaj}{cb3441@columbia.edu}

\icmlkeywords{Machine Learning, ICML, Deep Learning, SGLD, SGD, Optimization, Noisy Gradient, Noisy Optimization}

\vskip 0.3in
]



\printAffiliationsAndNotice{}  

\begin{abstract}

Stochastic Gradient Langevin Dynamics infuses isotropic gradient noise to SGD to help navigate pathological curvature in the loss landscape for deep networks. Isotropic nature of the noise leads to poor scaling, and adaptive methods based on higher order curvature information such as Fisher Scoring have been proposed to precondition the noise in order to achieve better convergence. In this paper, we describe an adaptive method to estimate the parameters of the noise and conduct experiments on well-known model architectures to show that the adaptively preconditioned SGLD method achieves convergence with the speed of adaptive first order methods such as Adam, AdaGrad etc. and achieves generalization equivalent of SGD in the test set. 
\end{abstract}

\section{Introduction}
\label{Introduction}

Generalizability is the ability of a model to perform well on unseen examples \cite{gp}. Neural networks are know to overfit the data, and mechanisms such as regularization are employed to constrain a model's ability to learn in order to reduce the generalization gap.

Various schemes such as dropout \cite{dropout}, weight decay \cite{weight} and early stopping \cite{earlystop1,earlystop2,earlystop3} have been proposed to regularize neural network models. Regularization in neural networks can be roughly categorized into implicit methods \cite{implicit} and explicit methods \cite{overp}. The ability of Stochastic Gradient Descent (SGD) to generalize better than other adaptive optimization methods is often attributed to its role as an implicit regularization mechanism.
 
Various adaptive optimization methods such as RMSProp \cite{rmsprop}, Adam \cite{adam}, AdaGrad \cite{adagrad} and AMSGrad \cite{amsgrad} have been proposed to speed up the training of deep networks. First order adaptive methods typically have a faster training speed, but Stochastic Gradient Descent  is often found to achieve better generalization on the test set \cite{hyp,adabound}.

 Stochastic Gradient Langevin Dynamics (SGLD) \citep{sgld} adds an isotropic noise to SGD to help it navigate out of saddle points and suboptimal local minima. SGLD has a powerful Bayesian interpretation, and is often used in Monte Carlo Markov Chains to sample the posterior for inference \cite{mandt}. 
 
The slow convergence of SGD while training is due to the uniform scaling in the parameter space. Adaptive methods conventionally speed up training by applying an element wise scaling scheme. Various approaches to pre-condition the noise in SGLD on the basis of higher order information such as Fisher Scoring \cite{sgfs} have been shown to achieve better generalizability than SGD, but such higher order methods have a high computational complexity and hence not scalable to very deep networks.
 
In this paper, we propose a method to adaptively estimate the parameters of noise in SGLD using first order information in order to achieve high training speed and better generalizability.

\section{Related Work}
Adding noise to the input, the model structure or the gradient updates itself is a well-studied topic \cite{oldnoise}. The success of mini-batch gradient descent over batch gradient descent is attributed to the variance brought due to constraint in the sampling procedure.
 	
Methods such as weight noise \cite{weightnoise} and adaptive weight noise \cite{adaptiveweightg,adaptiveweightb} infuses noise by perturbing the weights with a Gaussian Noise. Dropout randomly drops neurons with a probability, and it mimics training an ensemble of neural networks.
 
Hamiltonian Monte Carlo \cite{hmc1,hmc2} works by sampling a posterior using noise to explore the state space. A mini-batch variant of Hamiltonian Monte Carlo is Stochastic Gradient Langevin Dynamics (SGLD). The noise in SGLD help it better explore the loss landscape and also helps it navigate out of malformed curvature such as saddle points and sub-optimal local minima. 
	
Various adaptive optimization algorithms have been proposed to improve the speed of training of neural networks such as Adam, AdaGrad and AMSGrad. The adaptive methods apply an element wise scaling on the gradients to allow for faster convergence. The adaptive algorithms perform incredibly well for convex settings, but are not able to generalize as well as SGD for non-convex problems.
 
    Similar to SGD, the slow convergence in SGLD is attributed to uniform scaling in the parameter space. The speed of convergence and generalizability of the method can be improved using an adaptive preconditioner on the noise.  
 
   Scaling of noise in SGLD can be performed by using a pre-conditioner \cite{psgld}. Second order pre-conditioners encoding inverse Hessians \cite{12} and Fisher Information \cite{fisherinfo,kfac} have been used to establish better generalizability but suffer from high computational complexity. First order methods based on RMSProp \cite{psgld} use the second order moments of the gradients to inversely scale the noise, thereby increasing noise in sensitive dimensions and dampening noise in dimensions with large gradients.

We propose a method to scale the noise proportionaly to the second order moment of the gradients in order to achieve a higher training speed by increasing the noise for dimensions with larger gradients. In this paper we describe a method to create a pre-conditioner for SGLD which possess the training speed of adaptive methods and the generalizability of SGD with minimal computational overhead.

\section{Adaptively Preconditioned SGLD}

Consider a supervised learning problem, where we have identically distributed data and label pairs \( (x_1,y_1),.., (x_n,y_n) \in \mathbb{R}^{d+1} \). Our goal is to optimize the distribution \( p(y|x) \) by minimizing an approximate loss function \( \mathcal{L} (y_i| x_i, \Theta) \) with respect to \( \Theta \), where the distribution \( p(y|x) \) is parametrized by \( \Theta \).

Finding the optimal parameters for a Neural Network is a known NP-hard problem \cite{implicit,nphard}. The parameters of a probability distribution occupy a Riemannian Manifold \cite{amari}, and greedy optimization methods such as Stochastic Gradient Descent exploit curvature information in the manifold to find the most optimal parameters of the distribution in a convex case. Stochastic Gradient Descent optimizes the loss function using gradients of the loss function with respect to the parameter at each step

\begin{equation}
\hat{g}_{s}(\Theta_t) \leftarrow \nabla_{\Theta} \hat{\mathcal{L}}_{s}(\Theta_t)
\end{equation}

where \( \hat{\mathcal{L}}_{s}(\Theta) \) is the stochastic estimate of the loss function computed over a mini-batch of size \(s \) sampled uniformly from the data. The parameter updates can be written as

\begin{equation}
\Theta_{t+1} \leftarrow \Theta_t - \eta (\hat{g}_{s}(\Theta_t))
\end{equation}

SGD with decreasing step sizes provably converges to the optimum of a convex function, and to the local optimum in case of a non-convex function \cite{monro}.

The loss landscape of very deep neural networks is often ill-behaved and non-convex in nature. To navigate out of sub-optimal local minima, strategies such as momentum \cite{momentum1,momentum2} are employed

\begin{equation}
\mu_{t} \leftarrow \rho \mu_{t-1} + (1 - \rho)\hat{g}_{s}(\Theta_t)  
\end{equation}

\begin{equation}
\Theta_{t+1} \leftarrow \Theta_t - \eta (\mu_{t})
\end{equation}

\begin{algorithm}[tb]
   \caption{Adaptively Preconditioned SGLD}
   \label{alg:example}
\begin{algorithmic}
   \STATE Input \( : \Theta_0 \), step size \( \eta \), momentum \(\rho\), noise \(\psi \)
   \STATE Set \( \mu_0 =0 \) and \( \sigma_0=0 \)
   \FOR{ t = 1 to T}
   		\STATE \( \hat{g}_{s}(\Theta_t) \leftarrow \nabla_{\Theta} \hat{\mathcal{L}}_{s}(\Theta_t) \)
   		\STATE \( \mu_{t} \leftarrow \rho \mu_{t-1} + (1 - \rho)\hat{g}_{s}(\Theta_t)  \)

   		\STATE \( C_{t} \leftarrow \rho C_{t-1}+ (1-\rho)(\hat{g}_{s}(\Theta_t)- \mu_t )(\hat{g}_{s}(\Theta_t)- \mu_{t-1} ) \)
   		
   		\STATE \( \xi_t \sim  N(\mu_t,C_t) \)
   		
   		\STATE \( \Theta_{t+1} \leftarrow \Theta_t- \eta ( \hat{g}_{s}(\Theta_t) + \psi \xi_t) \)
   \ENDFOR
\end{algorithmic}
\end{algorithm}

\begin{figure}[t]
\centering
\vskip 0.2in
\subfigure[ASGLD vs SGD (with momentum)]{\label{fig:a}\includegraphics[width=\columnwidth]{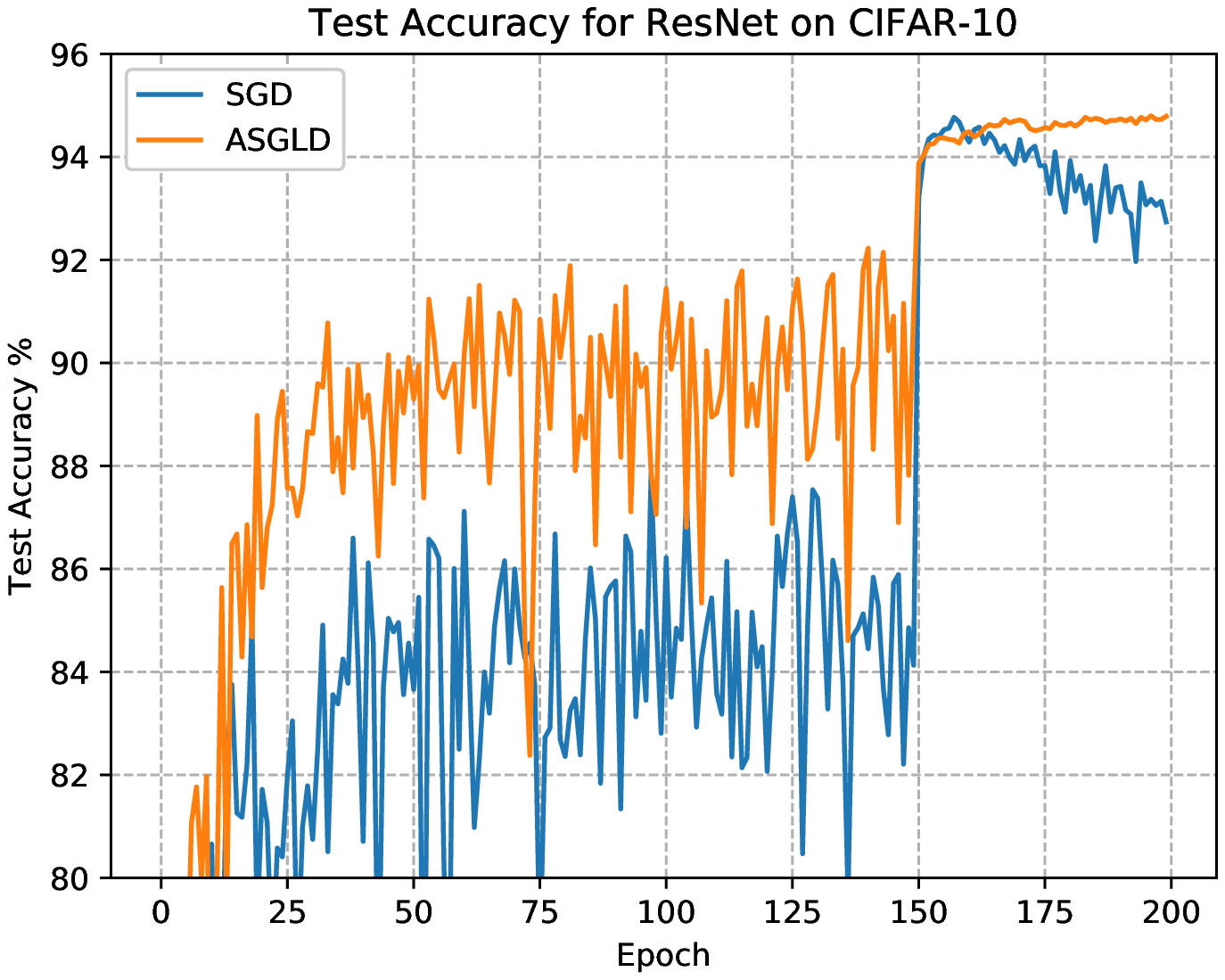}} \quad
\vskip -0.2in
\vskip 0.2in

\subfigure[ASGLD vs adaptive methods]{\label{fig:b}\includegraphics[width=\columnwidth]{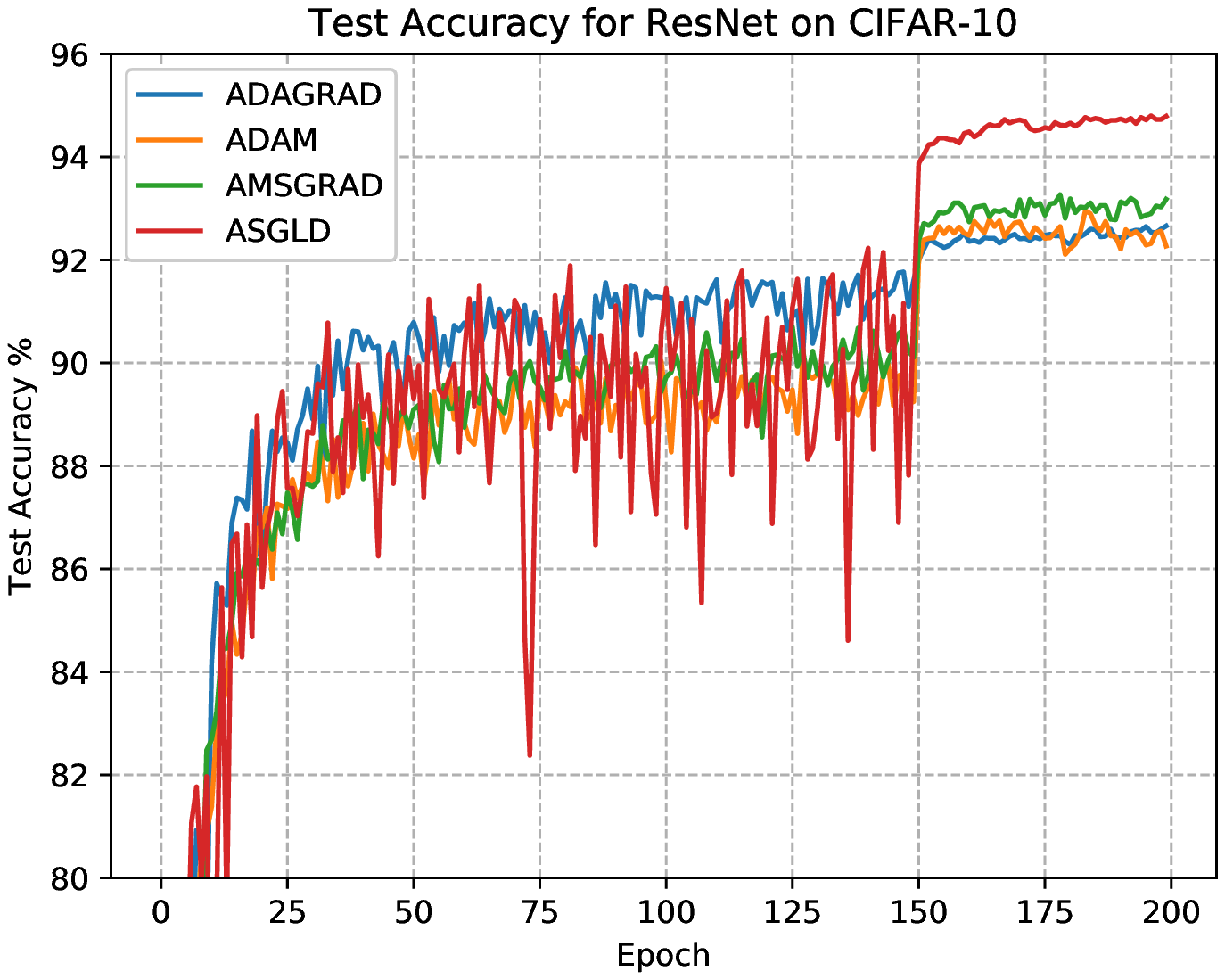}}
\caption{Comparison of ASGLD vs conventional optimization methods on CIFAR 10 with Resnet 34 architecture}
\vskip -0.2in

\end{figure}

\begin{figure}[t]
\centering
\vskip 0.2in
\subfigure[ASGLD vs SGD (with momentum)]{\label{fig:a}\includegraphics[width=\columnwidth]{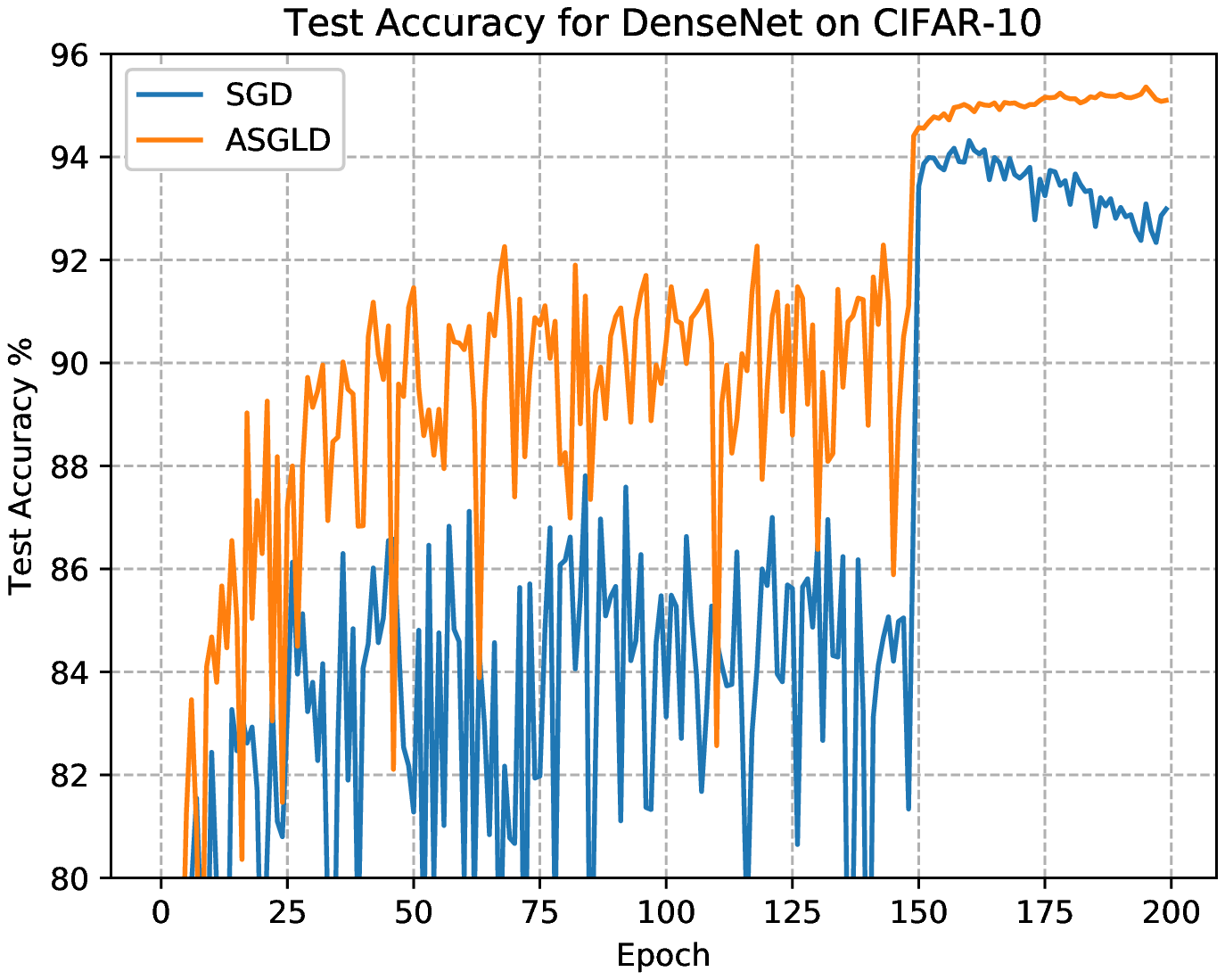}}
\vskip -0.2in
\vskip 0.2in
\subfigure[ASGLD vs adaptive methods]{\label{fig:b}\includegraphics[width=\columnwidth]{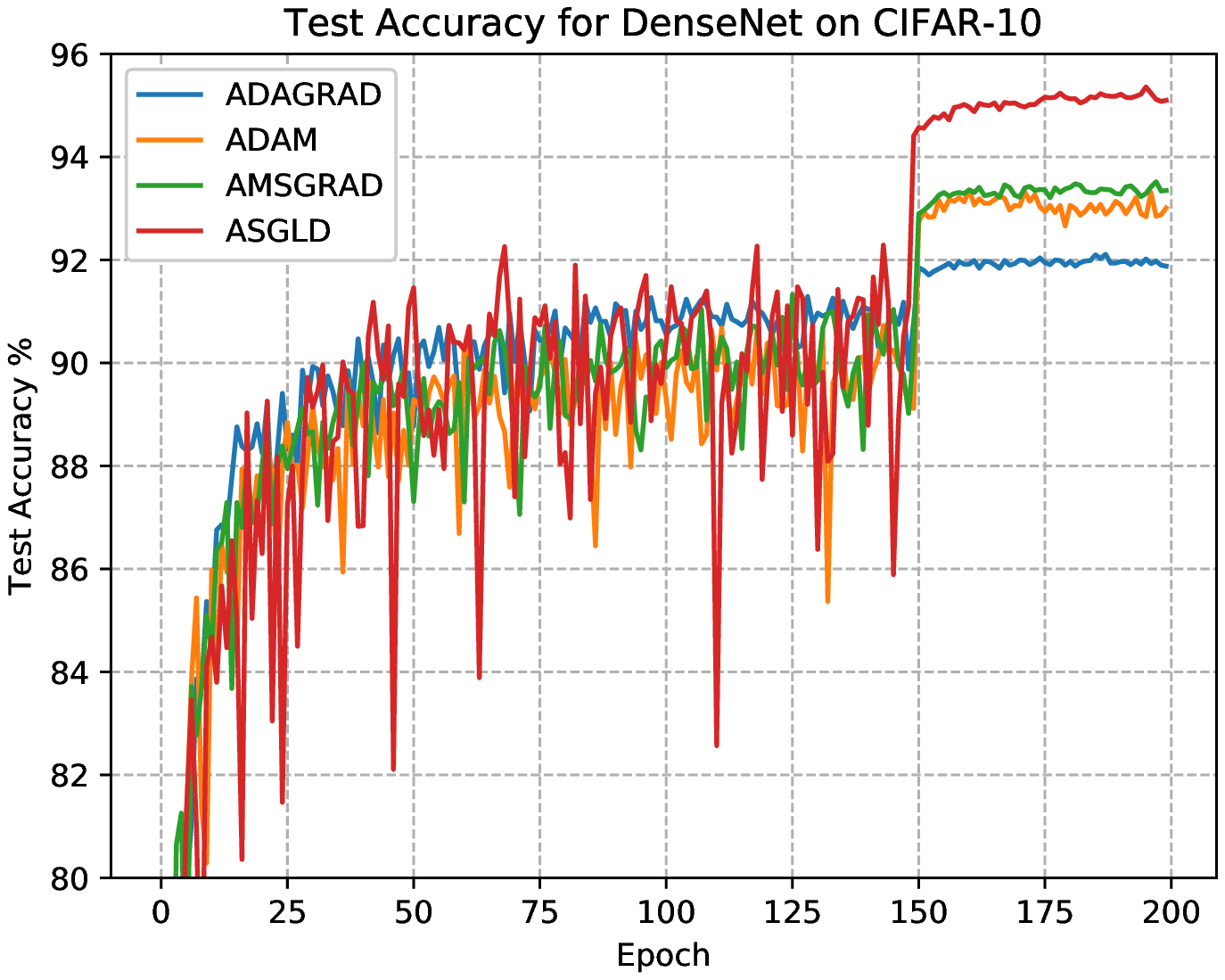}}
\caption{Comparison of ASGLD vs conventional methods on CIFAR 10 with Densenet 121 architecture}
\vskip -0.2in

\end{figure}

Stochastic Gradient Langevin Dynamics (SGLD) further extends SGD by adding additional Gaussian noise to help it escape sub-optimal minima. We can approximate SGLD using 

\begin{equation}
\xi_t \sim N(0,\epsilon)
\end{equation}

\begin{equation}
\Theta_{t+1} \leftarrow \Theta_t- \eta ( \hat{g}_{s}(\Theta_t) + \xi_t )
\end{equation}

SGLD can also be provably shown to converge to the optimal minima in a convex case when limit \( \epsilon , \eta \rightarrow 0 \) holds. \cite{mandt}. Stochastic Gradient Hamiltonian Monte Carlo Stochastic Gradient \cite{sghmc} adds momentum to SGLD 

\begin{equation}
\Theta_{t+1} \leftarrow \Theta_t- \eta ( \mu_t + \xi_t )
\end{equation}

The equi-scaled nature of noise leads to poor scaling of parameter updates, leading to a slower training speed and risk of converging to a sub-optimal minima \cite{adabound}.  Noise can be adaptively pre-conditioned to help traverse pathological curvature

\begin{equation}
\xi_t \sim  N(0,C)
\end{equation}

Preconditioners based on higher order information use the inverse of Hessian or Fisher Information matrix to help traverse the curvature better. Unfortunately, such higher order approaches are computationally infeasible for large and deep networks. Adaptive pre-conditioners based on popular adaptive methods such as RMSProp use a diagonal approximation of the inverse of second order moments of the gradient updates.

Adaptive pre-conditioning methods yield similar or better generalization performance versus SGD, but still possess a rather slower speed of convergence with respect to adaptive first order methods \cite{sgldc}.

We propose an adaptive preconditioner based on a diagonal approximation of second order moment of gradient updates, which posses the generalizability of SGD and the training speed of adaptive first order methods. Adaptively Preconditioned SGLD (ASGLD) method scales the noise in a directly proportional manner to allow for faster training speed

\begin{equation}
C_{t} \leftarrow \rho C_{t-1}+ (1-\rho)(\hat{g}_{s}(\Theta_t)- \mu_t )(\hat{g}_{s}(\Theta_t)- \mu_{t-1} )
\end{equation}

\begin{equation}
\xi_t \sim  N(\mu_t,C_t)
\end{equation}

\begin{equation}
\Theta_{t+1} \leftarrow \Theta_t- \eta ( \hat{g}_{s}(\Theta_t) + \psi \xi_t)
\end{equation}
where $\psi$ is the noise parameter.

The noise covariance preconditioner scales the noise proportionally in dimensions with larger gradients, essentially helping it escape suboptimal minima and saddle points better and thus helping it converge faster and to a better solution. As the algorithm approaches a wide minima, the dampened second order moment starts shrinking, allowing for convergence to the optimum.

\section{Experiments}

In this section, we examine the impact of using ASGLD method on Resnet 34 \cite{resnet} and Densenet 121 \cite{densenet} architectures on CIFAR 10 dataset \cite{cifar}. CIFAR 10 dataset contains 60,000 images for ten classes sampled from tiny images dataset.

  Training was performed for a fixed schedule of 200 runs over the training set, and we plot the training and test accuracy in fig 1. and fig 2. We reduce the learning rate by a factor of 10 at the 150th epoch.
  
   We performed hyperparameter tuning in accordance with methods defined in \cite{hyp} and \cite{adabound}. For learning rate tuning, we implement a logarithmically spaced grid with five step sizes, and we try new grid points if the best performing parameter setting is found at one end of the grid. We match the settings for other hyperparameters such as batch size, weight decay and dropout probability with the respective base architectures.

In Resenet 34 architecture, we observe in fig 1.a) that ASGLD performs better in terms of training speed than SGD and achieves similar accuracy on the held out set. We also observe in fig 1.b) that ASGLD has similar training speed as first order adaptive methods early in training, but ASGLD begins to significantly outperform adaptive methods in generalization error by the time the learning rates are decayed. We also observe that ASGLD is more stable as compared to SGD at the end of the training.  

We see similar trends for Densenet 121 architecture, as evident in fig 2. We also observe that ASGLD has a lower generalization error than SGD at the end of the training.

\section{Discussion}

 To investigate the ability of our method, we conducted experiments on CIFAR 10 using well known neural network architectures such as Resnet 34 and Densenet 121. Based on the results obtained in the experiment section, we can observe that ASGLD performs as well as adaptive methods and much better than SGD early in training, but by the time the learning rate is decayed we observe that the ASGLD method performs as well as SGD and significantly outperforms first order adaptive methods. 

 Furthermore, we also observed similar wall clock times for training with the ASGLD method versus adaptive methods.

\section{Future Work}

Future work would include exploring the impact of ASGLD on other regularization mechanisms such as Batch Normalization etc. 

We would also like to investigate the effectiveness of ASGLD on other domains such as Natural Language Processing, Speech etc.

\section{Conclusion}

We propose a new method ASGLD based on adaptively preconditioning noise covariance matrix in SGLD using estimated second order moments of gradient updates for optimizing a non-convex function, and demonstrate its effectiveness over well-known datasets using popular neural network architectures.

We observe that ASGLD method significantly outperforms adaptive methods in generalizability and SGD in terms of speed of convergence and stability. We also observe the increased effectiveness of ASGLD in deeper networks. 

\section{Acknowledgment}

We thank Zoran Kostic for providing valuable comments and computing resources.

\bibliography{ASGLD}

\begin{thebibliography}{37}
\providecommand{\natexlab}[1]{#1}
\providecommand{\url}[1]{\texttt{#1}}
\expandafter\ifx\csname urlstyle\endcsname\relax
  \providecommand{\doi}[1]{doi: #1}\else
  \providecommand{\doi}{doi: \begingroup \urlstyle{rm}\Url}\fi

\bibitem[Ahn et~al.(2012)Ahn, Korattikara, and Welling]{sgfs}
Ahn, S., Korattikara, A., and Welling, M.
\newblock Bayesian posterior sampling via stochastic gradient fisher scoring.
\newblock \emph{arXiv preprint arXiv:1206.6380}, 2012.

\bibitem[Allen-Zhu(2018)]{nphard}
Allen-Zhu, Z.
\newblock Natasha 2: Faster non-convex optimization than sgd.
\newblock In \emph{Advances in Neural Information Processing Systems}, pp.\
  2675--2686, 2018.

\bibitem[Amari(1998)]{amari}
Amari, S.-I.
\newblock Natural gradient works efficiently in learning.
\newblock \emph{Neural computation}, 10\penalty0 (2):\penalty0 251--276, 1998.

\bibitem[An(1996)]{oldnoise}
An, G.
\newblock The effects of adding noise during backpropagation training on a
  generalization performance.
\newblock \emph{Neural computation}, 8\penalty0 (3):\penalty0 643--674, 1996.

\bibitem[Blundell et~al.(2015)Blundell, Cornebise, Kavukcuoglu, and
  Wierstra]{adaptiveweightb}
Blundell, C., Cornebise, J., Kavukcuoglu, K., and Wierstra, D.
\newblock Weight uncertainty in neural networks.
\newblock \emph{arXiv preprint arXiv:1505.05424}, 2015.

\bibitem[Caruana et~al.(2001)Caruana, Lawrence, and Giles]{earlystop2}
Caruana, R., Lawrence, S., and Giles, C.~L.
\newblock Overfitting in neural nets: Backpropagation, conjugate gradient, and
  early stopping.
\newblock In \emph{Advances in neural information processing systems}, pp.\
  402--408, 2001.

\bibitem[Chen et~al.(2014)Chen, Fox, and Guestrin]{sghmc}
Chen, T., Fox, E., and Guestrin, C.
\newblock Stochastic gradient hamiltonian monte carlo.
\newblock In \emph{International conference on machine learning}, pp.\
  1683--1691, 2014.

\bibitem[Duane et~al.(1987)Duane, Kennedy, Pendleton, and Roweth]{hmc1}
Duane, S., Kennedy, A.~D., Pendleton, B.~J., and Roweth, D.
\newblock Hybrid monte carlo.
\newblock \emph{Physics letters B}, 195\penalty0 (2):\penalty0 216--222, 1987.

\bibitem[Duchi et~al.(2011)Duchi, Hazan, and Singer]{adagrad}
Duchi, J., Hazan, E., and Singer, Y.
\newblock Adaptive subgradient methods for online learning and stochastic
  optimization.
\newblock \emph{Journal of Machine Learning Research}, 12\penalty0
  (Jul):\penalty0 2121--2159, 2011.

\bibitem[Graves(2011)]{adaptiveweightg}
Graves, A.
\newblock Practical variational inference for neural networks.
\newblock In \emph{Advances in neural information processing systems}, pp.\
  2348--2356, 2011.

\bibitem[He et~al.(2016)He, Zhang, Ren, and Sun]{resnet}
He, K., Zhang, X., Ren, S., and Sun, J.
\newblock Deep residual learning for image recognition.
\newblock In \emph{Proceedings of the IEEE conference on computer vision and
  pattern recognition}, pp.\  770--778, 2016.

\bibitem[Huang et~al.(2017)Huang, Liu, Van Der~Maaten, and
  Weinberger]{densenet}
Huang, G., Liu, Z., Van Der~Maaten, L., and Weinberger, K.~Q.
\newblock Densely connected convolutional networks.
\newblock In \emph{Proceedings of the IEEE conference on computer vision and
  pattern recognition}, pp.\  4700--4708, 2017.

\bibitem[Jiang et~al.(2019)Jiang, Krishnan, Mobahi, and Bengio]{gp}
Jiang, Y., Krishnan, D., Mobahi, H., and Bengio, S.
\newblock A margin-based measure of generalization for deep networks.
\newblock 2019.

\bibitem[Kingma \& Ba(2014)Kingma and Ba]{adam}
Kingma, D.~P. and Ba, J.
\newblock Adam: A method for stochastic optimization.
\newblock \emph{arXiv preprint arXiv:1412.6980}, 2014.

\bibitem[Krizhevsky et~al.(2014)Krizhevsky, Nair, and Hinton]{cifar}
Krizhevsky, A., Nair, V., and Hinton, G.
\newblock The cifar-10 dataset.
\newblock \emph{online: http://www. cs. toronto. edu/kriz/cifar. html}, 55,
  2014.

\bibitem[Krogh \& Hertz(1992)Krogh and Hertz]{weight}
Krogh, A. and Hertz, J.~A.
\newblock A simple weight decay can improve generalization.
\newblock In \emph{Advances in neural information processing systems}, pp.\
  950--957, 1992.

\bibitem[Li et~al.(2016)Li, Chen, Carlson, and Carin]{psgld}
Li, C., Chen, C., Carlson, D., and Carin, L.
\newblock Preconditioned stochastic gradient langevin dynamics for deep neural
  networks.
\newblock In \emph{Thirtieth AAAI Conference on Artificial Intelligence}, 2016.

\bibitem[Luo et~al.(2019)Luo, Xiong, Liu, and Sun]{adabound}
Luo, L., Xiong, Y., Liu, Y., and Sun, X.
\newblock Adaptive gradient methods with dynamic bound of learning rate.
\newblock \emph{arXiv preprint arXiv:1902.09843}, 2019.

\bibitem[Mandt et~al.(2017)Mandt, Hoffman, and Blei]{mandt}
Mandt, S., Hoffman, M.~D., and Blei, D.~M.
\newblock Stochastic gradient descent as approximate bayesian inference.
\newblock \emph{The Journal of Machine Learning Research}, 18\penalty0
  (1):\penalty0 4873--4907, 2017.

\bibitem[Marceau-Caron \& Ollivier(2017)Marceau-Caron and Ollivier]{fisherinfo}
Marceau-Caron, G. and Ollivier, Y.
\newblock Natural langevin dynamics for neural networks.
\newblock In \emph{International Conference on Geometric Science of
  Information}, pp.\  451--459. Springer, 2017.

\bibitem[Martin et~al.(2012)Martin, Wilcox, Burstedde, and Ghattas]{12}
Martin, J., Wilcox, L.~C., Burstedde, C., and Ghattas, O.
\newblock A stochastic newton mcmc method for large-scale statistical inverse
  problems with application to seismic inversion.
\newblock \emph{SIAM Journal on Scientific Computing}, 34\penalty0
  (3):\penalty0 A1460--A1487, 2012.

\bibitem[Nado et~al.(2018)Nado, Snoek, Grosse, Duvenaud, Xu, and Martens]{kfac}
Nado, Z., Snoek, J., Grosse, R., Duvenaud, D., Xu, B., and Martens, J.
\newblock Stochastic gradient langevin dynamics that exploit neural network
  structure.
\newblock 2018.

\bibitem[Neal et~al.(2011)]{hmc2}
Neal, R.~M. et~al.
\newblock Mcmc using hamiltonian dynamics.
\newblock \emph{Handbook of markov chain monte carlo}, 2\penalty0
  (11):\penalty0 2, 2011.

\bibitem[Neyshabur et~al.(2014)Neyshabur, Tomioka, and Srebro]{implicit}
Neyshabur, B., Tomioka, R., and Srebro, N.
\newblock In search of the real inductive bias: On the role of implicit
  regularization in deep learning.
\newblock \emph{arXiv preprint arXiv:1412.6614}, 2014.

\bibitem[Neyshabur et~al.(2018)Neyshabur, Li, Bhojanapalli, LeCun, and
  Srebro]{overp}
Neyshabur, B., Li, Z., Bhojanapalli, S., LeCun, Y., and Srebro, N.
\newblock The role of over-parametrization in generalization of neural
  networks.
\newblock 2018.

\bibitem[Palacci \& Hess(2018)Palacci and Hess]{sgldc}
Palacci, H. and Hess, H.
\newblock Scalable natural gradient langevin dynamics in practice.
\newblock \emph{arXiv preprint arXiv:1806.02855}, 2018.

\bibitem[Polyak(1964)]{momentum1}
Polyak, B.~T.
\newblock Some methods of speeding up the convergence of iteration methods.
\newblock \emph{USSR Computational Mathematics and Mathematical Physics},
  4\penalty0 (5):\penalty0 1--17, 1964.

\bibitem[Prechelt(1998)]{earlystop1}
Prechelt, L.
\newblock Automatic early stopping using cross validation: quantifying the
  criteria.
\newblock \emph{Neural Networks}, 11\penalty0 (4):\penalty0 761--767, 1998.

\bibitem[Reddi et~al.(2019)Reddi, Kale, and Kumar]{amsgrad}
Reddi, S.~J., Kale, S., and Kumar, S.
\newblock On the convergence of adam and beyond.
\newblock \emph{arXiv preprint arXiv:1904.09237}, 2019.

\bibitem[Robbins \& Monro(1951)Robbins and Monro]{monro}
Robbins, H. and Monro, S.
\newblock A stochastic approximation method.
\newblock \emph{The annals of mathematical statistics}, pp.\  400--407, 1951.

\bibitem[Srivastava et~al.(2014)Srivastava, Hinton, Krizhevsky, Sutskever, and
  Salakhutdinov]{dropout}
Srivastava, N., Hinton, G., Krizhevsky, A., Sutskever, I., and Salakhutdinov,
  R.
\newblock Dropout: a simple way to prevent neural networks from overfitting.
\newblock \emph{The Journal of Machine Learning Research}, 15\penalty0
  (1):\penalty0 1929--1958, 2014.

\bibitem[Steijvers \& Gr{\"u}nwald(1996)Steijvers and
  Gr{\"u}nwald]{weightnoise}
Steijvers, M. and Gr{\"u}nwald, P.
\newblock A recurrent network that performs a context-sensitive prediction
  task.
\newblock In \emph{Proceedings of the 18th annual conference of the cognitive
  science society}, pp.\  335--339, 1996.

\bibitem[Sutskever et~al.(2013)Sutskever, Martens, Dahl, and Hinton]{momentum2}
Sutskever, I., Martens, J., Dahl, G., and Hinton, G.
\newblock On the importance of initialization and momentum in deep learning.
\newblock In \emph{International conference on machine learning}, pp.\
  1139--1147, 2013.

\bibitem[Tieleman \& Hinton(2012)Tieleman and Hinton]{rmsprop}
Tieleman, T. and Hinton, G.
\newblock Lecture 6.5-rmsprop: Divide the gradient by a running average of its
  recent magnitude.
\newblock \emph{COURSERA: Neural networks for machine learning}, 4\penalty0
  (2):\penalty0 26--31, 2012.

\bibitem[Welling \& Teh(2011)Welling and Teh]{sgld}
Welling, M. and Teh, Y.~W.
\newblock Bayesian learning via stochastic gradient langevin dynamics.
\newblock In \emph{Proceedings of the 28th international conference on machine
  learning (ICML-11)}, pp.\  681--688, 2011.

\bibitem[Wilson et~al.(2017)Wilson, Roelofs, Stern, Srebro, and Recht]{hyp}
Wilson, A.~C., Roelofs, R., Stern, M., Srebro, N., and Recht, B.
\newblock The marginal value of adaptive gradient methods in machine learning.
\newblock In \emph{Advances in Neural Information Processing Systems}, pp.\
  4148--4158, 2017.

\bibitem[Yao et~al.(2007)Yao, Rosasco, and Caponnetto]{earlystop3}
Yao, Y., Rosasco, L., and Caponnetto, A.
\newblock On early stopping in gradient descent learning.
\newblock \emph{Constructive Approximation}, 26\penalty0 (2):\penalty0
  289--315, 2007.

\end{thebibliography}
\bibliographystyle{icml2019}

\end{document}